\documentclass[runningheads]{llncs}
\usepackage{graphicx}
\usepackage{amsmath,amssymb} % define this before the line numbering.
\usepackage{color}
\usepackage[draft]{minted}

\def\PYGdefault@toks#1+{\ifx\relax#1\empty\else%
    \PYGdefault@tok{#1}\expandafter\PYGdefault@toks\fi}
\def\PYGdefault@do#1{\PYGdefault@bc{\PYGdefault@tc{\PYGdefault@ul{%
    \PYGdefault@it{\PYGdefault@bf{\PYGdefault@ff{#1}}}}}}}
\def\PYGdefault#1#2{\PYGdefault@reset\PYGdefault@toks#1+\relax+\PYGdefault@do{#2}}

\expandafter\def\csname PYGdefault@tok@gd\endcsname{\def\PYGdefault@tc##1{\textcolor[rgb]{0.63,0.00,0.00}{##1}}}
\expandafter\def\csname PYGdefault@tok@gu\endcsname{\let\PYGdefault@bf=\textbf\def\PYGdefault@tc##1{\textcolor[rgb]{0.50,0.00,0.50}{##1}}}
\expandafter\def\csname PYGdefault@tok@gt\endcsname{\def\PYGdefault@tc##1{\textcolor[rgb]{0.00,0.27,0.87}{##1}}}
\expandafter\def\csname PYGdefault@tok@gs\endcsname{\let\PYGdefault@bf=\textbf}
\expandafter\def\csname PYGdefault@tok@gr\endcsname{\def\PYGdefault@tc##1{\textcolor[rgb]{1.00,0.00,0.00}{##1}}}
\expandafter\def\csname PYGdefault@tok@cm\endcsname{\let\PYGdefault@it=\textit\def\PYGdefault@tc##1{\textcolor[rgb]{0.25,0.50,0.50}{##1}}}
\expandafter\def\csname PYGdefault@tok@vg\endcsname{\def\PYGdefault@tc##1{\textcolor[rgb]{0.10,0.09,0.49}{##1}}}
\expandafter\def\csname PYGdefault@tok@m\endcsname{\def\PYGdefault@tc##1{\textcolor[rgb]{0.40,0.40,0.40}{##1}}}
\expandafter\def\csname PYGdefault@tok@mh\endcsname{\def\PYGdefault@tc##1{\textcolor[rgb]{0.40,0.40,0.40}{##1}}}
\expandafter\def\csname PYGdefault@tok@go\endcsname{\def\PYGdefault@tc##1{\textcolor[rgb]{0.53,0.53,0.53}{##1}}}
\expandafter\def\csname PYGdefault@tok@ge\endcsname{\let\PYGdefault@it=\textit}
\expandafter\def\csname PYGdefault@tok@vc\endcsname{\def\PYGdefault@tc##1{\textcolor[rgb]{0.10,0.09,0.49}{##1}}}
\expandafter\def\csname PYGdefault@tok@il\endcsname{\def\PYGdefault@tc##1{\textcolor[rgb]{0.40,0.40,0.40}{##1}}}
\expandafter\def\csname PYGdefault@tok@cs\endcsname{\let\PYGdefault@it=\textit\def\PYGdefault@tc##1{\textcolor[rgb]{0.25,0.50,0.50}{##1}}}
\expandafter\def\csname PYGdefault@tok@cp\endcsname{\def\PYGdefault@tc##1{\textcolor[rgb]{0.74,0.48,0.00}{##1}}}
\expandafter\def\csname PYGdefault@tok@gi\endcsname{\def\PYGdefault@tc##1{\textcolor[rgb]{0.00,0.63,0.00}{##1}}}
\expandafter\def\csname PYGdefault@tok@gh\endcsname{\let\PYGdefault@bf=\textbf\def\PYGdefault@tc##1{\textcolor[rgb]{0.00,0.00,0.50}{##1}}}
\expandafter\def\csname PYGdefault@tok@ni\endcsname{\let\PYGdefault@bf=\textbf\def\PYGdefault@tc##1{\textcolor[rgb]{0.60,0.60,0.60}{##1}}}
\expandafter\def\csname PYGdefault@tok@nl\endcsname{\def\PYGdefault@tc##1{\textcolor[rgb]{0.63,0.63,0.00}{##1}}}
\expandafter\def\csname PYGdefault@tok@nn\endcsname{\let\PYGdefault@bf=\textbf\def\PYGdefault@tc##1{\textcolor[rgb]{0.00,0.00,1.00}{##1}}}
\expandafter\def\csname PYGdefault@tok@no\endcsname{\def\PYGdefault@tc##1{\textcolor[rgb]{0.53,0.00,0.00}{##1}}}
\expandafter\def\csname PYGdefault@tok@na\endcsname{\def\PYGdefault@tc##1{\textcolor[rgb]{0.49,0.56,0.16}{##1}}}
\expandafter\def\csname PYGdefault@tok@nb\endcsname{\def\PYGdefault@tc##1{\textcolor[rgb]{0.00,0.50,0.00}{##1}}}
\expandafter\def\csname PYGdefault@tok@nc\endcsname{\let\PYGdefault@bf=\textbf\def\PYGdefault@tc##1{\textcolor[rgb]{0.00,0.00,1.00}{##1}}}
\expandafter\def\csname PYGdefault@tok@nd\endcsname{\def\PYGdefault@tc##1{\textcolor[rgb]{0.67,0.13,1.00}{##1}}}
\expandafter\def\csname PYGdefault@tok@ne\endcsname{\let\PYGdefault@bf=\textbf\def\PYGdefault@tc##1{\textcolor[rgb]{0.82,0.25,0.23}{##1}}}
\expandafter\def\csname PYGdefault@tok@nf\endcsname{\def\PYGdefault@tc##1{\textcolor[rgb]{0.00,0.00,1.00}{##1}}}
\expandafter\def\csname PYGdefault@tok@si\endcsname{\let\PYGdefault@bf=\textbf\def\PYGdefault@tc##1{\textcolor[rgb]{0.73,0.40,0.53}{##1}}}
\expandafter\def\csname PYGdefault@tok@s2\endcsname{\def\PYGdefault@tc##1{\textcolor[rgb]{0.73,0.13,0.13}{##1}}}
\expandafter\def\csname PYGdefault@tok@vi\endcsname{\def\PYGdefault@tc##1{\textcolor[rgb]{0.10,0.09,0.49}{##1}}}
\expandafter\def\csname PYGdefault@tok@nt\endcsname{\let\PYGdefault@bf=\textbf\def\PYGdefault@tc##1{\textcolor[rgb]{0.00,0.50,0.00}{##1}}}
\expandafter\def\csname PYGdefault@tok@nv\endcsname{\def\PYGdefault@tc##1{\textcolor[rgb]{0.10,0.09,0.49}{##1}}}
\expandafter\def\csname PYGdefault@tok@s1\endcsname{\def\PYGdefault@tc##1{\textcolor[rgb]{0.73,0.13,0.13}{##1}}}
\expandafter\def\csname PYGdefault@tok@kd\endcsname{\let\PYGdefault@bf=\textbf\def\PYGdefault@tc##1{\textcolor[rgb]{0.00,0.50,0.00}{##1}}}
\expandafter\def\csname PYGdefault@tok@sh\endcsname{\def\PYGdefault@tc##1{\textcolor[rgb]{0.73,0.13,0.13}{##1}}}
\expandafter\def\csname PYGdefault@tok@sc\endcsname{\def\PYGdefault@tc##1{\textcolor[rgb]{0.73,0.13,0.13}{##1}}}
\expandafter\def\csname PYGdefault@tok@sx\endcsname{\def\PYGdefault@tc##1{\textcolor[rgb]{0.00,0.50,0.00}{##1}}}
\expandafter\def\csname PYGdefault@tok@bp\endcsname{\def\PYGdefault@tc##1{\textcolor[rgb]{0.00,0.50,0.00}{##1}}}
\expandafter\def\csname PYGdefault@tok@c1\endcsname{\let\PYGdefault@it=\textit\def\PYGdefault@tc##1{\textcolor[rgb]{0.25,0.50,0.50}{##1}}}
\expandafter\def\csname PYGdefault@tok@kc\endcsname{\let\PYGdefault@bf=\textbf\def\PYGdefault@tc##1{\textcolor[rgb]{0.00,0.50,0.00}{##1}}}
\expandafter\def\csname PYGdefault@tok@c\endcsname{\let\PYGdefault@it=\textit\def\PYGdefault@tc##1{\textcolor[rgb]{0.25,0.50,0.50}{##1}}}
\expandafter\def\csname PYGdefault@tok@mf\endcsname{\def\PYGdefault@tc##1{\textcolor[rgb]{0.40,0.40,0.40}{##1}}}
\expandafter\def\csname PYGdefault@tok@err\endcsname{\def\PYGdefault@bc##1{\setlength{\fboxsep}{0pt}\fcolorbox[rgb]{1.00,0.00,0.00}{1,1,1}{\strut ##1}}}
\expandafter\def\csname PYGdefault@tok@mb\endcsname{\def\PYGdefault@tc##1{\textcolor[rgb]{0.40,0.40,0.40}{##1}}}
\expandafter\def\csname PYGdefault@tok@ss\endcsname{\def\PYGdefault@tc##1{\textcolor[rgb]{0.10,0.09,0.49}{##1}}}
\expandafter\def\csname PYGdefault@tok@sr\endcsname{\def\PYGdefault@tc##1{\textcolor[rgb]{0.73,0.40,0.53}{##1}}}
\expandafter\def\csname PYGdefault@tok@mo\endcsname{\def\PYGdefault@tc##1{\textcolor[rgb]{0.40,0.40,0.40}{##1}}}
\expandafter\def\csname PYGdefault@tok@kn\endcsname{\let\PYGdefault@bf=\textbf\def\PYGdefault@tc##1{\textcolor[rgb]{0.00,0.50,0.00}{##1}}}
\expandafter\def\csname PYGdefault@tok@mi\endcsname{\def\PYGdefault@tc##1{\textcolor[rgb]{0.40,0.40,0.40}{##1}}}
\expandafter\def\csname PYGdefault@tok@gp\endcsname{\let\PYGdefault@bf=\textbf\def\PYGdefault@tc##1{\textcolor[rgb]{0.00,0.00,0.50}{##1}}}
\expandafter\def\csname PYGdefault@tok@o\endcsname{\def\PYGdefault@tc##1{\textcolor[rgb]{0.40,0.40,0.40}{##1}}}
\expandafter\def\csname PYGdefault@tok@kr\endcsname{\let\PYGdefault@bf=\textbf\def\PYGdefault@tc##1{\textcolor[rgb]{0.00,0.50,0.00}{##1}}}
\expandafter\def\csname PYGdefault@tok@s\endcsname{\def\PYGdefault@tc##1{\textcolor[rgb]{0.73,0.13,0.13}{##1}}}
\expandafter\def\csname PYGdefault@tok@kp\endcsname{\def\PYGdefault@tc##1{\textcolor[rgb]{0.00,0.50,0.00}{##1}}}
\expandafter\def\csname PYGdefault@tok@w\endcsname{\def\PYGdefault@tc##1{\textcolor[rgb]{0.73,0.73,0.73}{##1}}}
\expandafter\def\csname PYGdefault@tok@kt\endcsname{\def\PYGdefault@tc##1{\textcolor[rgb]{0.69,0.00,0.25}{##1}}}
\expandafter\def\csname PYGdefault@tok@ow\endcsname{\let\PYGdefault@bf=\textbf\def\PYGdefault@tc##1{\textcolor[rgb]{0.67,0.13,1.00}{##1}}}
\expandafter\def\csname PYGdefault@tok@sb\endcsname{\def\PYGdefault@tc##1{\textcolor[rgb]{0.73,0.13,0.13}{##1}}}
\expandafter\def\csname PYGdefault@tok@k\endcsname{\let\PYGdefault@bf=\textbf\def\PYGdefault@tc##1{\textcolor[rgb]{0.00,0.50,0.00}{##1}}}
\expandafter\def\csname PYGdefault@tok@se\endcsname{\let\PYGdefault@bf=\textbf\def\PYGdefault@tc##1{\textcolor[rgb]{0.73,0.40,0.13}{##1}}}
\expandafter\def\csname PYGdefault@tok@sd\endcsname{\let\PYGdefault@it=\textit\def\PYGdefault@tc##1{\textcolor[rgb]{0.73,0.13,0.13}{##1}}}

\makeatother

\makeatletter
\def\PYG@reset{\let\PYG@it=\relax \let\PYG@bf=\relax%
    \let\PYG@ul=\relax \let\PYG@tc=\relax%
    \let\PYG@bc=\relax \let\PYG@ff=\relax}
\def\PYG@tok#1{\csname PYG@tok@#1\endcsname}
\def\PYG@toks#1+{\ifx\relax#1\empty\else%
    \PYG@tok{#1}\expandafter\PYG@toks\fi}
\def\PYG@do#1{\PYG@bc{\PYG@tc{\PYG@ul{%
    \PYG@it{\PYG@bf{\PYG@ff{#1}}}}}}}
\def\PYG#1#2{\PYG@reset\PYG@toks#1+\relax+\PYG@do{#2}}

\expandafter\def\csname PYG@tok@gd\endcsname{\def\PYG@tc##1{\textcolor[rgb]{0.63,0.00,0.00}{##1}}}
\expandafter\def\csname PYG@tok@gu\endcsname{\let\PYG@bf=\textbf\def\PYG@tc##1{\textcolor[rgb]{0.50,0.00,0.50}{##1}}}
\expandafter\def\csname PYG@tok@gt\endcsname{\def\PYG@tc##1{\textcolor[rgb]{0.00,0.27,0.87}{##1}}}
\expandafter\def\csname PYG@tok@gs\endcsname{\let\PYG@bf=\textbf}
\expandafter\def\csname PYG@tok@gr\endcsname{\def\PYG@tc##1{\textcolor[rgb]{1.00,0.00,0.00}{##1}}}
\expandafter\def\csname PYG@tok@cm\endcsname{\let\PYG@it=\textit\def\PYG@tc##1{\textcolor[rgb]{0.25,0.50,0.50}{##1}}}
\expandafter\def\csname PYG@tok@vg\endcsname{\def\PYG@tc##1{\textcolor[rgb]{0.10,0.09,0.49}{##1}}}
\expandafter\def\csname PYG@tok@m\endcsname{\def\PYG@tc##1{\textcolor[rgb]{0.40,0.40,0.40}{##1}}}
\expandafter\def\csname PYG@tok@mh\endcsname{\def\PYG@tc##1{\textcolor[rgb]{0.40,0.40,0.40}{##1}}}
\expandafter\def\csname PYG@tok@go\endcsname{\def\PYG@tc##1{\textcolor[rgb]{0.53,0.53,0.53}{##1}}}
\expandafter\def\csname PYG@tok@ge\endcsname{\let\PYG@it=\textit}
\expandafter\def\csname PYG@tok@vc\endcsname{\def\PYG@tc##1{\textcolor[rgb]{0.10,0.09,0.49}{##1}}}
\expandafter\def\csname PYG@tok@il\endcsname{\def\PYG@tc##1{\textcolor[rgb]{0.40,0.40,0.40}{##1}}}
\expandafter\def\csname PYG@tok@cs\endcsname{\let\PYG@it=\textit\def\PYG@tc##1{\textcolor[rgb]{0.25,0.50,0.50}{##1}}}
\expandafter\def\csname PYG@tok@cp\endcsname{\def\PYG@tc##1{\textcolor[rgb]{0.74,0.48,0.00}{##1}}}
\expandafter\def\csname PYG@tok@gi\endcsname{\def\PYG@tc##1{\textcolor[rgb]{0.00,0.63,0.00}{##1}}}
\expandafter\def\csname PYG@tok@gh\endcsname{\let\PYG@bf=\textbf\def\PYG@tc##1{\textcolor[rgb]{0.00,0.00,0.50}{##1}}}
\expandafter\def\csname PYG@tok@ni\endcsname{\let\PYG@bf=\textbf\def\PYG@tc##1{\textcolor[rgb]{0.60,0.60,0.60}{##1}}}
\expandafter\def\csname PYG@tok@nl\endcsname{\def\PYG@tc##1{\textcolor[rgb]{0.63,0.63,0.00}{##1}}}
\expandafter\def\csname PYG@tok@nn\endcsname{\let\PYG@bf=\textbf\def\PYG@tc##1{\textcolor[rgb]{0.00,0.00,1.00}{##1}}}
\expandafter\def\csname PYG@tok@no\endcsname{\def\PYG@tc##1{\textcolor[rgb]{0.53,0.00,0.00}{##1}}}
\expandafter\def\csname PYG@tok@na\endcsname{\def\PYG@tc##1{\textcolor[rgb]{0.49,0.56,0.16}{##1}}}
\expandafter\def\csname PYG@tok@nb\endcsname{\def\PYG@tc##1{\textcolor[rgb]{0.00,0.50,0.00}{##1}}}
\expandafter\def\csname PYG@tok@nc\endcsname{\let\PYG@bf=\textbf\def\PYG@tc##1{\textcolor[rgb]{0.00,0.00,1.00}{##1}}}
\expandafter\def\csname PYG@tok@nd\endcsname{\def\PYG@tc##1{\textcolor[rgb]{0.67,0.13,1.00}{##1}}}
\expandafter\def\csname PYG@tok@ne\endcsname{\let\PYG@bf=\textbf\def\PYG@tc##1{\textcolor[rgb]{0.82,0.25,0.23}{##1}}}
\expandafter\def\csname PYG@tok@nf\endcsname{\def\PYG@tc##1{\textcolor[rgb]{0.00,0.00,1.00}{##1}}}
\expandafter\def\csname PYG@tok@si\endcsname{\let\PYG@bf=\textbf\def\PYG@tc##1{\textcolor[rgb]{0.73,0.40,0.53}{##1}}}
\expandafter\def\csname PYG@tok@s2\endcsname{\def\PYG@tc##1{\textcolor[rgb]{0.73,0.13,0.13}{##1}}}
\expandafter\def\csname PYG@tok@vi\endcsname{\def\PYG@tc##1{\textcolor[rgb]{0.10,0.09,0.49}{##1}}}
\expandafter\def\csname PYG@tok@nt\endcsname{\let\PYG@bf=\textbf\def\PYG@tc##1{\textcolor[rgb]{0.00,0.50,0.00}{##1}}}
\expandafter\def\csname PYG@tok@nv\endcsname{\def\PYG@tc##1{\textcolor[rgb]{0.10,0.09,0.49}{##1}}}
\expandafter\def\csname PYG@tok@s1\endcsname{\def\PYG@tc##1{\textcolor[rgb]{0.73,0.13,0.13}{##1}}}
\expandafter\def\csname PYG@tok@kd\endcsname{\let\PYG@bf=\textbf\def\PYG@tc##1{\textcolor[rgb]{0.00,0.50,0.00}{##1}}}
\expandafter\def\csname PYG@tok@sh\endcsname{\def\PYG@tc##1{\textcolor[rgb]{0.73,0.13,0.13}{##1}}}
\expandafter\def\csname PYG@tok@sc\endcsname{\def\PYG@tc##1{\textcolor[rgb]{0.73,0.13,0.13}{##1}}}
\expandafter\def\csname PYG@tok@sx\endcsname{\def\PYG@tc##1{\textcolor[rgb]{0.00,0.50,0.00}{##1}}}
\expandafter\def\csname PYG@tok@bp\endcsname{\def\PYG@tc##1{\textcolor[rgb]{0.00,0.50,0.00}{##1}}}
\expandafter\def\csname PYG@tok@c1\endcsname{\let\PYG@it=\textit\def\PYG@tc##1{\textcolor[rgb]{0.25,0.50,0.50}{##1}}}
\expandafter\def\csname PYG@tok@kc\endcsname{\let\PYG@bf=\textbf\def\PYG@tc##1{\textcolor[rgb]{0.00,0.50,0.00}{##1}}}
\expandafter\def\csname PYG@tok@c\endcsname{\let\PYG@it=\textit\def\PYG@tc##1{\textcolor[rgb]{0.25,0.50,0.50}{##1}}}
\expandafter\def\csname PYG@tok@mf\endcsname{\def\PYG@tc##1{\textcolor[rgb]{0.40,0.40,0.40}{##1}}}
\expandafter\def\csname PYG@tok@err\endcsname{\def\PYG@bc##1{\setlength{\fboxsep}{0pt}\fcolorbox[rgb]{1.00,0.00,0.00}{1,1,1}{\strut ##1}}}
\expandafter\def\csname PYG@tok@mb\endcsname{\def\PYG@tc##1{\textcolor[rgb]{0.40,0.40,0.40}{##1}}}
\expandafter\def\csname PYG@tok@ss\endcsname{\def\PYG@tc##1{\textcolor[rgb]{0.10,0.09,0.49}{##1}}}
\expandafter\def\csname PYG@tok@sr\endcsname{\def\PYG@tc##1{\textcolor[rgb]{0.73,0.40,0.53}{##1}}}
\expandafter\def\csname PYG@tok@mo\endcsname{\def\PYG@tc##1{\textcolor[rgb]{0.40,0.40,0.40}{##1}}}
\expandafter\def\csname PYG@tok@kn\endcsname{\let\PYG@bf=\textbf\def\PYG@tc##1{\textcolor[rgb]{0.00,0.50,0.00}{##1}}}
\expandafter\def\csname PYG@tok@mi\endcsname{\def\PYG@tc##1{\textcolor[rgb]{0.40,0.40,0.40}{##1}}}
\expandafter\def\csname PYG@tok@gp\endcsname{\let\PYG@bf=\textbf\def\PYG@tc##1{\textcolor[rgb]{0.00,0.00,0.50}{##1}}}
\expandafter\def\csname PYG@tok@o\endcsname{\def\PYG@tc##1{\textcolor[rgb]{0.40,0.40,0.40}{##1}}}
\expandafter\def\csname PYG@tok@kr\endcsname{\let\PYG@bf=\textbf\def\PYG@tc##1{\textcolor[rgb]{0.00,0.50,0.00}{##1}}}
\expandafter\def\csname PYG@tok@s\endcsname{\def\PYG@tc##1{\textcolor[rgb]{0.73,0.13,0.13}{##1}}}
\expandafter\def\csname PYG@tok@kp\endcsname{\def\PYG@tc##1{\textcolor[rgb]{0.00,0.50,0.00}{##1}}}
\expandafter\def\csname PYG@tok@w\endcsname{\def\PYG@tc##1{\textcolor[rgb]{0.73,0.73,0.73}{##1}}}
\expandafter\def\csname PYG@tok@kt\endcsname{\def\PYG@tc##1{\textcolor[rgb]{0.69,0.00,0.25}{##1}}}
\expandafter\def\csname PYG@tok@ow\endcsname{\let\PYG@bf=\textbf\def\PYG@tc##1{\textcolor[rgb]{0.67,0.13,1.00}{##1}}}
\expandafter\def\csname PYG@tok@sb\endcsname{\def\PYG@tc##1{\textcolor[rgb]{0.73,0.13,0.13}{##1}}}
\expandafter\def\csname PYG@tok@k\endcsname{\let\PYG@bf=\textbf\def\PYG@tc##1{\textcolor[rgb]{0.00,0.50,0.00}{##1}}}
\expandafter\def\csname PYG@tok@se\endcsname{\let\PYG@bf=\textbf\def\PYG@tc##1{\textcolor[rgb]{0.73,0.40,0.13}{##1}}}
\expandafter\def\csname PYG@tok@sd\endcsname{\let\PYG@it=\textit\def\PYG@tc##1{\textcolor[rgb]{0.73,0.13,0.13}{##1}}}

\def\PYGZbs{\char`\\}
\def\PYGZus{\char`\_}
\def\PYGZob{\char`\{}
\def\PYGZcb{\char`\}}
\def\PYGZca{\char`\^}
\def\PYGZam{\char`\&}
\def\PYGZlt{\char`\<}
\def\PYGZgt{\char`\>}
\def\PYGZsh{\char`\#}
\def\PYGZpc{\char`\%}
\def\PYGZdl{\char`\$}
\def\PYGZhy{\char`\-}
\def\PYGZsq{\char`\'}
\def\PYGZdq{\char`\"}
\def\PYGZti{\char`\~}
\def\PYGZat{@}
\def\PYGZlb{[}
\def\PYGZrb{]}
\makeatother

\usepackage{array}
\newcolumntype{L}[1]{>{\raggedright\let\newline\\\arraybackslash\hspace{0pt}}m{#1}}
\newcolumntype{C}[1]{>{\centering\let\newline\\\arraybackslash\hspace{0pt}}m{#1}}
\newcolumntype{R}[1]{>{\raggedleft\let\newline\\\arraybackslash\hspace{0pt}}m{#1}}

\usepackage[width=122mm,left=12mm,paperwidth=146mm,height=193mm,top=12mm,paperheight=217mm]{geometry}
\begin{document}

% \linenumbers
\pagestyle{headings}
\mainmatter

\title{gvnn: Neural Network Library for Geometric Computer Vision} % Replace with your title

\titlerunning{gvnn: Neural Network Library for Geometric Computer Vision}

\authorrunning{Handa \textit{et al.}}

\author{Ankur Handa\inst{1}, Michael Bloesch\inst{3}, Viorica P{\u a}tr{\u a}ucean\inst{2}, Simon Stent\inst{2}, John McCormac\inst{1}, Andrew Davison\inst{1} 
\\
{\small handa.ankur@gmail.com, bloeschm@ethz.ch, \{vp344,sais2\}@cam.ac.uk, \{brendon.mccormac13, ajd\}@ic.ac.uk}
}

\institute{
Dyson Robotics Laboratory,
Department of Computing,
Imperial College London
\and 
Department of Engineering, University of Cambridge
\and
Robotic Systems Lab, ETH Zurich
}

\maketitle

\begin{abstract}
We introduce \textbf{gvnn}, a neural network library in Torch aimed towards bridging the gap between classic geometric computer vision and deep learning. Inspired by the recent success of Spatial Transformer Networks, we propose several new layers which are often used as parametric transformations on the data in geometric computer vision. These layers can be inserted within a neural network much in the spirit of the original spatial transformers and allow backpropagation to enable end-to-end learning of a network involving any domain knowledge in geometric computer vision. This opens up applications in learning invariance to 3D geometric transformation for place recognition, end-to-end visual odometry, depth estimation and unsupervised learning through warping with a parametric transformation for image reconstruction error.
 
%geared towards geometric computer vision within a deep learning framework. Inspired by the success of Spatial Transformer Networks and their ability to bring together domain knowledge into a neural network, we propose several new layers that are often used in geometric vision as fixed computational. 
%\dots
\end{abstract}

\section{Introduction}

Spatial transformers \cite{Jaderberg:etal:NIPS2015} represent a class of differentiable layers that can be inserted in a standard convolutional neural network architecture to enable invariance to certain geometric transformations on the input data and warping for reconstruction error \cite{Patraucean:etal:ICLRW15}. In this work, we build upon the 2D transformation layers originally proposed in the spatial transformer networks  \cite{Jaderberg:etal:NIPS2015} and provide various novel extensions that perform geometric transformations which are often used in geometric computer vision. These layers have mostly no internal parameters that need learning but allow backpropagation and can be inserted in a neural network for any fixed differentiable geometric operation to be performed on the data. This opens up an exciting new path to blend ideas from geometric computer vision into deep learning architectural designs allowing the exploitation of problem-specific domain knowledge.

Geometric computer vision has heavily relied on generative parametric models of inverse computer graphics to enable reasoning and understanding of real physical environments that provide rich observations in the form of images or video streams. These fundamentals and principles have been very well understood and form the backbone of large-scale point cloud reconstruction from multi-view image data, visual odometry, and image registration. In this work, we provide a comprehensive library that allows implementation of various image registration and reconstruction methods using these geometric transformation modules within the framework of convolutional neural networks. This means that certain elements in the classic geometric vision based methods that are hand-engineered can be replaced by a module that can be learnt end-to-end within a neural network. Our library is implemented in Torch \cite{Collobert:etal:NIPS2011} and builds upon the open source implementation of spatial transformer networks \cite{STNImplementation}.

\section{gvnn: Geometric Vision with Neural Networks}
We introduce \textbf{gvnn}, a Torch package dedicated to performing transformations that are often used in geometric computer vision applications within a neural network. These transformations are implemented as fixed differentiable computational blocks that can be inserted within a convolutional neural network and are useful for manipulating the input data as per the domain knowledge in geometric computer vision. We expand on various novel transformation layers below that form the core part of the library built on top of the open source implementation \cite{STNImplementation} of spatial transformer networks. 

Let us assume that $\mathcal{C}$ represents the cost function being optimised by the neural network. For a regression network it can take the following form \textit{e.g.} $\mathcal{C} = \frac{1}{2}||\mathbf{y}_{pred} - \mathbf{y}_{gt}||^{2}$ where $\mathbf{y}_{pred}$ is a prediction vector produced by the network and $\mathbf{y}_{gt}$ is the corresponding ground truth vector. This allows us to propagate derivatives from the loss function back to the input to any layer in the network.

\subsection{Global Transformations}
We begin by extending the 2D transformations introduced in the original spatial transformer networks (STN) to their 3D counterparts. These transformations encode the global movement of the whole image \textit{i.e.} the same transformation is applied to every pixel in the image or any 3D point in the world.

\subsubsection{SO3 Layer} Rotations in our network are represented by the so(3) vector (or $\mathfrak{so}(3)$ skew symmetric matrix), which is compact 3$\times$1 vector representation, $\mathbf{v} = (v_{1}, v_{2}, v_{3})^{T}$, and is turned into a rotation matrix via the SO3 exponential map, \textit{i.e.} $\mathsf{R}(\mathbf{v}) = \exp([\mathbf{v}]_{\times})$. The backpropagation derivatives for $\mathbf{v}$ can be conveniently written as \cite{Gallego:etal:ARXIV13}  
\begin{eqnarray}
\frac{\partial \mathcal{C}}{\partial \mathbf{v}} &=& \frac{\partial \mathcal{C}}{\partial \mathsf{R}(\mathbf{v})}\cdot\frac{\partial \mathsf{R}(\mathbf{v})}{\partial \mathbf{v}}
\end{eqnarray}
where 
\begin{eqnarray}
\frac{\partial \mathsf{R}(\mathbf{v})}{\partial v_{i}} &=& \frac{v_{i}[\mathbf{v}]_{\times} + [\mathbf{v} \times (\mathsf{I}-\mathsf{R})e_{i}]_{\times} }{||\mathbf{v}||^{2}} \mathsf{R}
\end{eqnarray}
$[\hspace{3mm}]_{\times}$ turns a 3$\times$1 vector to a skew-symmetric matrix and $\times$ is a cross product operation. $\mathsf{I}$ is the Identity 
matrix and $e_{i}$ is the $i^{th}$ column of the Identity matrix. We have also implemented different parameterisations \textit{e.g.} quaternions and Euler-angles for rotations as additional layers. Below we show the code-snippet that performs backpropagation on this layer.

\begin{footnotesize}
\begin{Verbatim}[commandchars=\\\{\}]
\PYG{k}{function} \PYG{n+nf}{RotationSO3}\PYG{p}{:}\PYG{n}{updateGradInput}\PYG{p}{(}\PYG{n}{\PYGZus{}tranformParams}\PYG{p}{,} \PYG{n}{\PYGZus{}gradParams}\PYG{p}{)}

  \PYG{c+c1}{\PYGZhy{}\PYGZhy{} \PYGZus{}transformParams are the input parameters i.e. so3 vector}
  \PYG{c+c1}{\PYGZhy{}\PYGZhy{} \PYGZus{}gradParams is the derivative of the cost function}
  \PYG{c+c1}{\PYGZhy{}\PYGZhy{} with respect to the rotation matrix}

  \PYG{c+c1}{\PYGZhy{}\PYGZhy{} gradInput is the derivative of cost}
  \PYG{c+c1}{\PYGZhy{}\PYGZhy{} function with respect to so3 vector}

  \PYG{k+kd}{local} \PYG{n}{tParams}\PYG{p}{,} \PYG{n}{gradParams}
  \PYG{n}{tParams} \PYG{o}{=} \PYG{n}{\PYGZus{}tranformParams}
  \PYG{n}{gradParams} \PYG{o}{=} \PYG{n}{\PYGZus{}gradParams}\PYG{p}{:}\PYG{n}{clone}\PYG{p}{()}

  \PYG{k+kd}{local} \PYG{n}{batchSize} \PYG{o}{=} \PYG{n}{tParams}\PYG{p}{:}\PYG{n}{size}\PYG{p}{(}\PYG{l+m+mi}{1}\PYG{p}{)}
  \PYG{n}{self}\PYG{p}{.}\PYG{n}{gradInput}\PYG{p}{:}\PYG{n}{resizeAs}\PYG{p}{(}\PYG{n}{tParams}\PYG{p}{)}

  \PYG{k+kd}{local} \PYG{n}{rotDerv} \PYG{o}{=} \PYG{n}{torch}\PYG{p}{.}\PYG{n}{zeros}\PYG{p}{(}\PYG{n}{batchSize}\PYG{p}{,} \PYG{l+m+mi}{3}\PYG{p}{,} \PYG{l+m+mi}{3}\PYG{p}{):}\PYG{n}{typeAs}\PYG{p}{(}\PYG{n}{tParams}\PYG{p}{)}
  \PYG{k+kd}{local} \PYG{n}{gradInputRotationParams} \PYG{o}{=} \PYG{n}{self}\PYG{p}{.}\PYG{n}{gradInput}\PYG{p}{:}\PYG{n}{narrow}\PYG{p}{(}\PYG{l+m+mi}{2}\PYG{p}{,}\PYG{l+m+mi}{1}\PYG{p}{,}\PYG{l+m+mi}{1}\PYG{p}{)}

  \PYG{c+c1}{\PYGZhy{}\PYGZhy{} take the derivative with respect to v1}
  \PYG{n}{rotDerv} \PYG{o}{=} \PYG{n}{dR\PYGZus{}by\PYGZus{}dvi}\PYG{p}{(}\PYG{n}{tParams}\PYG{p}{,}\PYG{n}{self}\PYG{p}{.}\PYG{n}{rotationOutput}\PYG{p}{,}\PYG{l+m+mi}{1}\PYG{p}{,} \PYG{n}{self}\PYG{p}{.}\PYG{n}{threshold}\PYG{p}{)}
  \PYG{k+kd}{local} \PYG{n}{selectGradParams} \PYG{o}{=} \PYG{n}{gradParams}\PYG{p}{:}\PYG{n}{narrow}\PYG{p}{(}\PYG{l+m+mi}{2}\PYG{p}{,}\PYG{l+m+mi}{1}\PYG{p}{,}\PYG{l+m+mi}{3}\PYG{p}{):}\PYG{n}{narrow}\PYG{p}{(}\PYG{l+m+mi}{3}\PYG{p}{,}\PYG{l+m+mi}{1}\PYG{p}{,}\PYG{l+m+mi}{3}\PYG{p}{)}
  \PYG{n}{gradRotParams}\PYG{p}{:}\PYG{n}{copy}\PYG{p}{(}\PYG{n}{torch}\PYG{p}{.}\PYG{n}{cmul}\PYG{p}{(}\PYG{n}{rotDerv}\PYG{p}{,}\PYG{n}{selectGradParams}\PYG{p}{):}\PYG{n}{sum}\PYG{p}{(}\PYG{l+m+mi}{2}\PYG{p}{):}\PYG{n}{sum}\PYG{p}{(}\PYG{l+m+mi}{3}\PYG{p}{))}

  \PYG{c+c1}{\PYGZhy{}\PYGZhy{} take the derivative with respect to v2}
  \PYG{n}{rotDerv} \PYG{o}{=} \PYG{n}{dR\PYGZus{}by\PYGZus{}dvi}\PYG{p}{(}\PYG{n}{tParams}\PYG{p}{,}\PYG{n}{self}\PYG{p}{.}\PYG{n}{rotationOutput}\PYG{p}{,}\PYG{l+m+mi}{2}\PYG{p}{,} \PYG{n}{self}\PYG{p}{.}\PYG{n}{threshold}\PYG{p}{)}
  \PYG{n}{gradRotParams} \PYG{o}{=} \PYG{n}{self}\PYG{p}{.}\PYG{n}{gradInput}\PYG{p}{:}\PYG{n}{narrow}\PYG{p}{(}\PYG{l+m+mi}{2}\PYG{p}{,}\PYG{l+m+mi}{2}\PYG{p}{,}\PYG{l+m+mi}{1}\PYG{p}{)}
  \PYG{n}{gradRotParams}\PYG{p}{:}\PYG{n}{copy}\PYG{p}{(}\PYG{n}{torch}\PYG{p}{.}\PYG{n}{cmul}\PYG{p}{(}\PYG{n}{rotDerv}\PYG{p}{,}\PYG{n}{selectGradParams}\PYG{p}{):}\PYG{n}{sum}\PYG{p}{(}\PYG{l+m+mi}{2}\PYG{p}{):}\PYG{n}{sum}\PYG{p}{(}\PYG{l+m+mi}{3}\PYG{p}{))}

  \PYG{c+c1}{\PYGZhy{}\PYGZhy{} take the derivative with respect to v3}
  \PYG{n}{rotDerv} \PYG{o}{=} \PYG{n}{dR\PYGZus{}by\PYGZus{}dvi}\PYG{p}{(}\PYG{n}{tParams}\PYG{p}{,}\PYG{n}{self}\PYG{p}{.}\PYG{n}{rotationOutput}\PYG{p}{,}\PYG{l+m+mi}{3}\PYG{p}{,} \PYG{n}{self}\PYG{p}{.}\PYG{n}{threshold}\PYG{p}{)}
  \PYG{n}{gradRotParams} \PYG{o}{=} \PYG{n}{self}\PYG{p}{.}\PYG{n}{gradInput}\PYG{p}{:}\PYG{n}{narrow}\PYG{p}{(}\PYG{l+m+mi}{2}\PYG{p}{,}\PYG{l+m+mi}{3}\PYG{p}{,}\PYG{l+m+mi}{1}\PYG{p}{)}
  \PYG{n}{gradRotParams}\PYG{p}{:}\PYG{n}{copy}\PYG{p}{(}\PYG{n}{torch}\PYG{p}{.}\PYG{n}{cmul}\PYG{p}{(}\PYG{n}{rotDerv}\PYG{p}{,}\PYG{n}{selectGradParams}\PYG{p}{):}\PYG{n}{sum}\PYG{p}{(}\PYG{l+m+mi}{2}\PYG{p}{):}\PYG{n}{sum}\PYG{p}{(}\PYG{l+m+mi}{3}\PYG{p}{))}

  \PYG{k}{return} \PYG{n}{self}\PYG{p}{.}\PYG{n}{gradInput}

\PYG{k}{end}
\end{Verbatim}
%\begin{minted}{lua}
%function RotationSO3:updateGradInput(_tranformParams, _gradParams)
%   
%  -- _transformParams are the input parameters i.e. so3 vector
%  -- _gradParams is the derivative of the cost function 
%  -- with respect to the rotation matrix
%  
%  -- gradInput is the derivative of cost 
%  -- function with respect to so3 vector   
% 
%  local tParams, gradParams
%  tParams = _tranformParams
%  gradParams = _gradParams:clone()
%
%  local batchSize = tParams:size(1)
%  self.gradInput:resizeAs(tParams)
%    
%  local rotDerv = torch.zeros(batchSize, 3, 3):typeAs(tParams)
%  local gradInputRotationParams = self.gradInput:narrow(2,1,1)
%  
%  -- take the derivative with respect to v1
%  rotDerv = dR_by_dvi(tParams,self.rotationOutput,1, self.threshold)
%  local selectGradParams = gradParams:narrow(2,1,3):narrow(3,1,3)
%  gradRotParams:copy(torch.cmul(rotDerv,selectGradParams):sum(2):sum(3))
%
%  -- take the derivative with respect to v2
%  rotDerv = dR_by_dvi(tParams,self.rotationOutput,2, self.threshold)
%  gradRotParams = self.gradInput:narrow(2,2,1)
%  gradRotParams:copy(torch.cmul(rotDerv,selectGradParams):sum(2):sum(3))
%
%  -- take the derivative with respect to v3
%  rotDerv = dR_by_dvi(tParams,self.rotationOutput,3, self.threshold)
%  gradRotParams = self.gradInput:narrow(2,3,1)
%  gradRotParams:copy(torch.cmul(rotDerv,selectGradParams):sum(2):sum(3))
%
%  return self.gradInput
%
%end
%\end{minted}
\end{footnotesize}

\subsubsection{SE3 Layer} The SE3 layer adds translations on top of the SO3 layer where translations are represented by a 3$\times$1 vector $\mathbf{t}$, and together they make up the 3$\times$4 transformation, \textit{i.e.} $\mathsf{T} = [\mathsf{R} | \mathbf{t}] \in $ SE3. 

\subsubsection{Sim3 Layer} Sim3 layer builds on top of the SE3 layer and has an extra scale factor $s$ to allow for any scale changes associated with the transformations $\mathsf{T} = \begin{bmatrix}
       s\mathsf{R} & \mathbf{t}           \\[0.3em]
       \mathsf{0} & 1  \\[0.3em]
     \end{bmatrix}$.

\subsubsection{3D Grid Generator} The 3D grid generator is an extension of the 2D grid generator proposed in the original STN. It takes additionally a depth map as input, to map the image pixels to corresponding 3D points in the world and transforms these points with $\mathsf{T}$ coming from the SE3 layer. Note that we have used a regular grid in this layer, but it is possible to extend this to the general case where the grid locations can also be learnt.

\subsubsection{Projection Layer} Projection layer maps the transformed 3D points, $\mathbf{p} = (u, v, w)^{T}$, onto 2D image plane using the focal lengths and the camera centre location. \textit{i.e.}
\begin{eqnarray}
\pi\left(
    \begin{array}{c}
      u \\
      v \\
      w
    \end{array}
  \right) &=& 
      \left(\begin{array}{c}
      f_x\frac{u}{w} + p_x \\
      \vspace{-2mm}\\
      f_y\frac{v}{w} + p_y
    \end{array}\right)
\end{eqnarray} where $f_x$ and $f_y$ represent the focal lengths of the camera along X and Y axes and $p_x$ and $p_y$ are the camera center locations. The backpropagation derivatives can be written as 
\begin{eqnarray} 
\frac{\partial C}{\partial \mathbf{p}} &=& \frac{\partial C}{\partial \pi(\mathbf{p})}\cdot \frac{\partial \pi(\mathbf{p})}{\partial \mathbf{p}}
\end{eqnarray}
where 
\begin{eqnarray}  
\frac{\partial \pi\left(
    \begin{array}{c}
      u \\
      v \\
      w
    \end{array}
  \right)}{\partial \left(\begin{array}{c}
      u \\
      v \\
      w
    \end{array} \right)} 
 &=& 
\left(
    \begin{array}{ccc}
      f_x\frac{1}{w} & 0 & -f_x\frac{u}{w^2}\\
      \vspace{-2mm} \\
       0 & f_y\frac{1}{w} & -f_y\frac{v}{w^2}
    \end{array}
  \right)   
\end{eqnarray}
In fact, if focal lengths are also involved in the optimisation, it is straightforward to include them in the network for any geometric camera calibration style optimisations. Note that special care must be taken to ensure that $w$ is not very small. Fortunately, in many geometric vision problems $w$ corresponds to the $z$-coordinate of a 3D point and is measured in metres --- when using Kinect or ASUS xtion cameras this happens to be always greater than 10 cm\footnote{We discovered that anything below than that the forward/backward gradient check fails}. 

\subsection{Per-pixel Transformations}
In many computer vision problems, particularly related to understanding dynamic scenes, it is often required to have per-pixel transformations to model the movements of the stimuli in the scene. In the following, we propose different layers for modelling per-pixel transformations for both RGB and RGB-D inputs.

\subsubsection{RGB based}
In the context of RGB data, the classic optic flow problem is a case of per-pixel transformation to model the movement of pixels across time. We implement both the well-known minimal parameterisation in the form of translation as well as more recently studied over-parameterised formulations that encapsulate the knowledge of scene geometry into the flow movement.

\paragraph{Mimimal parameterisation optic flow} In its minimal parameterisation, optic flow $(t_x, t_y)$ models the movement of pixels in the 2D image plane \textit{i.e.}
\begin{eqnarray} 
\left(
    \begin{array}{c}
      x^{\prime} \\
      y^{\prime} \\
    \end{array}
  \right) &=& 
\left(
    \begin{array}{c}
      x + t_x\\
      y + t_y
    \end{array} 
\right)   
\end{eqnarray}
This is the most well-known and studied parameterisation of optic flow in the literature and needs only 2 parameters per-pixel. In general, an extra smoothness penalty is imposed to ensure that the gradient of the flow varies smoothly across a pixel neighbourhood. Patraucean \textit{et al.} \cite{Patraucean:etal:ICLRW15} implement exactly this to model the optic flow and use Huber penalty for smoothness. We include this as a part of our library together with recent extensions with over-parameterised formulations.
 
\paragraph{Over-parameterised optic flow} Attempts to use the popular differential epipolar constraint \cite{Brooks:etal:JOSA1997} and the recent over-parameterised formulations of \cite{Nir:etal:IJCV2008} and \cite{Hornavcek:etal:ECCV2014} have shown that if knowledge about the scene geometry and motion can be used, it can greatly improve the flow estimates per-pixel. For instance, if the pixel lies on a planar surface, the motion of the pixel can be modelled by an affine transformation. Although \cite{Hornavcek:etal:ECCV2014} use a 9-DoF per-pixel transformation that includes the knowledge about the homography, we describe the affine parameterisation used in \cite{Nir:etal:IJCV2008}.
\begin{eqnarray} 
\left(
    \begin{array}{c}
      x^{\prime} \\
      y^{\prime} \\
      
    \end{array}
  \right) &=& 
\left(
    \begin{array}{ccc}
      a_0 & a_1 & a_2 \\
      \vspace{-2mm} \\
      a_3 & a_4 & a_5
    \end{array} 
  \right)\left(
    \begin{array}{c}
      x \\
      y \\
      1 \\
    \end{array}
  \right)   
\end{eqnarray}
It is interesting to note that popular 2-DoF translation optic flow describe earlier happens to be a special case of affine transformation.
\begin{eqnarray} 
\left(
    \begin{array}{c}
      x^{\prime} \\
      y^{\prime} \\
      
    \end{array}
  \right) &=& 
\left(
    \begin{array}{p{2.0mm}p{2.0mm}c}
      1 & 0 & t_x\\
      %\vspace{-1mm} \\
      0 & 1 & t_y
    \end{array} 
  \right)\left(
    \begin{array}{c}
      x \\
      y \\
      1 \\
    \end{array}
  \right)   
\end{eqnarray}
We provide implementations of 6-DoF affine transformation as well as SE(2) transformation per-pixel but extensions to 9-DoF paramterisation \cite{Hornavcek:etal:ECCV2014} are straightforward.

\begin{footnotesize}
\begin{Verbatim}[commandchars=\\\{\}]
\PYG{k}{function} \PYG{n+nf}{AffineOpticFlow}\PYG{p}{:}\PYG{n}{updateGradInput}\PYG{p}{(}\PYG{n}{\PYGZus{}PerPixelAffineParams}\PYG{p}{,} \PYG{n}{\PYGZus{}gradGrid}\PYG{p}{)}

   \PYG{k+kd}{local} \PYG{n}{batchsize} \PYG{o}{=} \PYG{n}{\PYGZus{}PerPixelAffineParams}\PYG{p}{:}\PYG{n}{size}\PYG{p}{(}\PYG{l+m+mi}{1}\PYG{p}{)}

   \PYG{n}{self}\PYG{p}{.}\PYG{n}{gradInput}\PYG{p}{:}\PYG{n}{resizeAs}\PYG{p}{(}\PYG{n}{\PYGZus{}PerPixelAffineParams}\PYG{p}{):}\PYG{n}{zero}\PYG{p}{()}

   \PYG{c+c1}{\PYGZhy{}\PYGZhy{} batchGrid is the regular 2D grid: B H W 2}
   \PYG{c+c1}{\PYGZhy{}\PYGZhy{} batches: B, height: H, width: W, channels: 2}

   \PYG{k+kd}{local} \PYG{n}{Lx\PYGZus{}x} \PYG{o}{=} \PYG{n}{torch}\PYG{p}{.}\PYG{n}{cmul}\PYG{p}{(}\PYG{n}{\PYGZus{}gradGrid}\PYG{p}{:}\PYG{n+nb}{select}\PYG{p}{(}\PYG{l+m+mi}{4}\PYG{p}{,}\PYG{l+m+mi}{1}\PYG{p}{),} \PYG{n}{self}\PYG{p}{.}\PYG{n}{batchGrid}\PYG{p}{:}\PYG{n+nb}{select}\PYG{p}{(}\PYG{l+m+mi}{4}\PYG{p}{,}\PYG{l+m+mi}{1}\PYG{p}{))}
   \PYG{k+kd}{local} \PYG{n}{Lx\PYGZus{}y} \PYG{o}{=} \PYG{n}{torch}\PYG{p}{.}\PYG{n}{cmul}\PYG{p}{(}\PYG{n}{\PYGZus{}gradGrid}\PYG{p}{:}\PYG{n+nb}{select}\PYG{p}{(}\PYG{l+m+mi}{4}\PYG{p}{,}\PYG{l+m+mi}{1}\PYG{p}{),} \PYG{n}{self}\PYG{p}{.}\PYG{n}{batchGrid}\PYG{p}{:}\PYG{n+nb}{select}\PYG{p}{(}\PYG{l+m+mi}{4}\PYG{p}{,}\PYG{l+m+mi}{2}\PYG{p}{))}

   \PYG{k+kd}{local} \PYG{n}{Ly\PYGZus{}x} \PYG{o}{=} \PYG{n}{torch}\PYG{p}{.}\PYG{n}{cmul}\PYG{p}{(}\PYG{n}{\PYGZus{}gradGrid}\PYG{p}{:}\PYG{n+nb}{select}\PYG{p}{(}\PYG{l+m+mi}{4}\PYG{p}{,}\PYG{l+m+mi}{2}\PYG{p}{),} \PYG{n}{self}\PYG{p}{.}\PYG{n}{batchGrid}\PYG{p}{:}\PYG{n+nb}{select}\PYG{p}{(}\PYG{l+m+mi}{4}\PYG{p}{,}\PYG{l+m+mi}{1}\PYG{p}{))}
   \PYG{k+kd}{local} \PYG{n}{Ly\PYGZus{}y} \PYG{o}{=} \PYG{n}{torch}\PYG{p}{.}\PYG{n}{cmul}\PYG{p}{(}\PYG{n}{\PYGZus{}gradGrid}\PYG{p}{:}\PYG{n+nb}{select}\PYG{p}{(}\PYG{l+m+mi}{4}\PYG{p}{,}\PYG{l+m+mi}{2}\PYG{p}{),} \PYG{n}{self}\PYG{p}{.}\PYG{n}{batchGrid}\PYG{p}{:}\PYG{n+nb}{select}\PYG{p}{(}\PYG{l+m+mi}{4}\PYG{p}{,}\PYG{l+m+mi}{2}\PYG{p}{))}

   \PYG{n}{self}\PYG{p}{.}\PYG{n}{gradInput}\PYG{p}{:}\PYG{n+nb}{select}\PYG{p}{(}\PYG{l+m+mi}{4}\PYG{p}{,}\PYG{l+m+mi}{1}\PYG{p}{):}\PYG{n}{copy}\PYG{p}{(}\PYG{n}{Lx\PYGZus{}x}\PYG{p}{)}
   \PYG{n}{self}\PYG{p}{.}\PYG{n}{gradInput}\PYG{p}{:}\PYG{n+nb}{select}\PYG{p}{(}\PYG{l+m+mi}{4}\PYG{p}{,}\PYG{l+m+mi}{2}\PYG{p}{):}\PYG{n}{copy}\PYG{p}{(}\PYG{n}{Lx\PYGZus{}y}\PYG{p}{)}
   \PYG{n}{self}\PYG{p}{.}\PYG{n}{gradInput}\PYG{p}{:}\PYG{n+nb}{select}\PYG{p}{(}\PYG{l+m+mi}{4}\PYG{p}{,}\PYG{l+m+mi}{3}\PYG{p}{):}\PYG{n}{copy}\PYG{p}{(}\PYG{n}{\PYGZus{}gradGrid}\PYG{p}{:}\PYG{n+nb}{select}\PYG{p}{(}\PYG{l+m+mi}{4}\PYG{p}{,}\PYG{l+m+mi}{1}\PYG{p}{))}

   \PYG{n}{self}\PYG{p}{.}\PYG{n}{gradInput}\PYG{p}{:}\PYG{n+nb}{select}\PYG{p}{(}\PYG{l+m+mi}{4}\PYG{p}{,}\PYG{l+m+mi}{4}\PYG{p}{):}\PYG{n}{copy}\PYG{p}{(}\PYG{n}{Ly\PYGZus{}x}\PYG{p}{)}
   \PYG{n}{self}\PYG{p}{.}\PYG{n}{gradInput}\PYG{p}{:}\PYG{n+nb}{select}\PYG{p}{(}\PYG{l+m+mi}{4}\PYG{p}{,}\PYG{l+m+mi}{5}\PYG{p}{):}\PYG{n}{copy}\PYG{p}{(}\PYG{n}{Ly\PYGZus{}y}\PYG{p}{)}
   \PYG{n}{self}\PYG{p}{.}\PYG{n}{gradInput}\PYG{p}{:}\PYG{n+nb}{select}\PYG{p}{(}\PYG{l+m+mi}{4}\PYG{p}{,}\PYG{l+m+mi}{6}\PYG{p}{):}\PYG{n}{copy}\PYG{p}{(}\PYG{n}{\PYGZus{}gradGrid}\PYG{p}{:}\PYG{n+nb}{select}\PYG{p}{(}\PYG{l+m+mi}{4}\PYG{p}{,}\PYG{l+m+mi}{2}\PYG{p}{))}

   \PYG{k}{return} \PYG{n}{self}\PYG{p}{.}\PYG{n}{gradInput}

\PYG{k}{end}
\end{Verbatim}
%\begin{minted}{lua}
%function AffineOpticFlow:updateGradInput(_PerPixelAffineParams, _gradGrid)
%
%   local batchsize = _PerPixelAffineParams:size(1)
%
%   self.gradInput:resizeAs(_PerPixelAffineParams):zero()
% 
%   -- batchGrid is the regular 2D grid: B H W 2
%   -- batches: B, height: H, width: W, channels: 2
%
%   local Lx_x = torch.cmul(_gradGrid:select(4,1), self.batchGrid:select(4,1))
%   local Lx_y = torch.cmul(_gradGrid:select(4,1), self.batchGrid:select(4,2))
%
%   local Ly_x = torch.cmul(_gradGrid:select(4,2), self.batchGrid:select(4,1))
%   local Ly_y = torch.cmul(_gradGrid:select(4,2), self.batchGrid:select(4,2))
%
%   self.gradInput:select(4,1):copy(Lx_x)
%   self.gradInput:select(4,2):copy(Lx_y)
%   self.gradInput:select(4,3):copy(_gradGrid:select(4,1))
%
%   self.gradInput:select(4,4):copy(Ly_x)
%   self.gradInput:select(4,5):copy(Ly_y)
%   self.gradInput:select(4,6):copy(_gradGrid:select(4,2))
%
%   return self.gradInput
%
%end
%\end{minted}
\end{footnotesize}

\paragraph{Slanted plane depth disparity} Similar ideas have been used in \cite{Bleyer:etal:2011} to obtain disparity of a stereo pair. They exploit the fact that scenes can be decomposed into piecewise slanted planes and consequently the disparity of a pixel can be expressed by the plane equation. This results in a over-paramterised 3-DoF formulation of disparity.
\begin{eqnarray}
d &=& ax + by + c
\end{eqnarray}
Again, this over-parameterisation greatly improves the results. Note that this formulation can be easily generalised and lifted to higher dimensions in the spirit of Total Generalised Variation (TGV) \cite{Pock:etal:2011}, but we have only implemented the 3-DoF formulation. 

We would like to stress that these layers are particularly tailored towards warping images which could be used as a direct signal for feedback loop in image reconstruction error in unsupervised training \cite{Patraucean:etal:ICLRW15,Garg:etal:arXiv2016}.

\subsubsection{RGB-D based}
Our layers can be easily adapted to RGB-D to enable 3D point cloud registration and alignment via per-pixel rigid transformations. Such transformations have been used extensively in the computer graphics community for some time and exploited by \cite{Sumner:etal:SIGGRAPH2007,Zollhofer:etal:TOG14,Newcombe:etal:CVPR2015} for non-rigid alignment. We extend similar ideas and implement 3D transformations for each pixel containing a 3D vector $\mathbf{x}$, the 3D spatial coordinates coming from a depth-map. In principle, such alignment is general and not limited to just 3D spatial points \textit{i.e.} any 3D feature per-pixel can be transformed. This is particularly useful when aligning feature maps as used in sketch and style transfer using deep learning \cite{Johnson:etal:2016}.

\paragraph{Per-pixel Sim3 transformation} We extend the global Sim3 transformation that models scale $s$, Rotation $\mathsf{R}$, and translation $\mathbf{t}$ to a per-pixel Sim3 transformation \textit{i.e.}  $\mathsf{T}_{i} = \begin{bmatrix}
       s_{i}\mathsf{R}_{i} & \mathbf{t}_{i}           \\[0.3em]
       \mathsf{0} & 1  \\[0.3em]
     \end{bmatrix}$ where $\mathsf{R} \in$ SO3.
\begin{eqnarray} 
\left(
    \begin{array}{c}
      x_i^{\prime} \\
      y_i^{\prime} \\
      z_i^{\prime}
    \end{array}
  \right) &=& 
    \mathsf{T}_i\left(
    \begin{array}{c}
      x_i \\
      y_i \\
      z_i \\
      1 \\
    \end{array}
  \right)   
\end{eqnarray}
This allows for the attention like mechanism of \cite{Jaderberg:etal:NIPS2015} in 3D, as specific voxel areas can be cropped and zoomed, and also modelling any 3D registrations that require scale. 
 
\paragraph{Per-pixel 10 DoF transformation} In many non-rigid alignments the rotation need not happen around the origin but around an anchor point $\mathbf{p}_i$ which is also jointly estimated. In this case, the transformation extends to 10 degrees of freedom \cite{Sumner:etal:SIGGRAPH2007}.
\begin{eqnarray} 
    \begin{array}{c}
      \mathbf{x}_{i}^{\prime}
    \end{array}
   &=& 
  \begin{array}{c}
    s_i(\mathsf{R}_i(\mathbf{x}_i - \mathbf{p}_i) + \mathbf{p}_i) + \mathbf{t}_i
    \end{array}
\end{eqnarray}
Additionally, smoothness constraints can be added to ensure that transformations are locally smooth in just the same way as Huber penalty is imposed for smoothing 2D optic flow. 

\begin{figure}[h]
\centerline{
\includegraphics[width=0.5\linewidth]{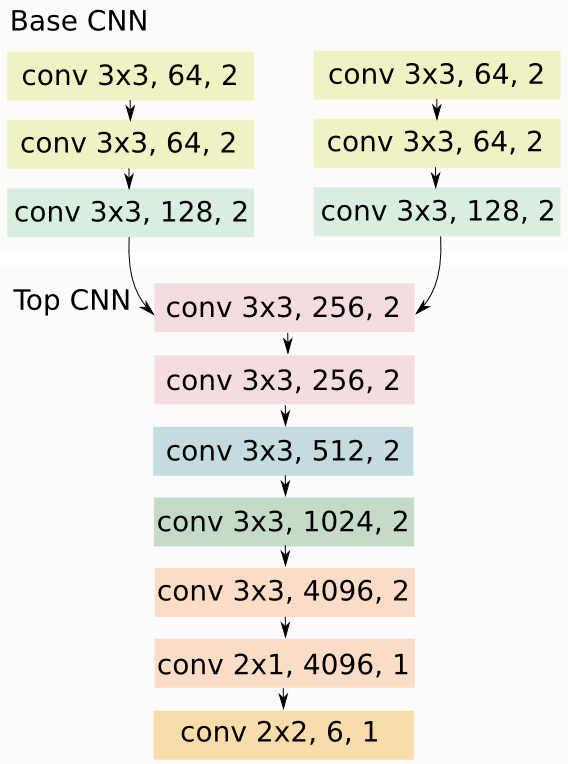}
}
\caption{\small
Our Siamese network is inspired by the popular VGG-16 network \cite{Simonyan:Zisserman:ICLR2015} where 3$\times$3 convolutions are used in most layers and works for 320$\times$240 image resolution. Each convolution layer is followed by PReLU non-linearity \cite{He:etal:ICCV2015}. We explicitly avoid any pooling and use a stride of 2 in every convolution layer for any downsampling.} 
\label{fig:Siamese Network}
\end{figure}

\subsection{M-estimators}
The standard least-squares loss function often employed in parameter fitting greatly affects the quality of the solution obtained at convergence. Built on the assumption that noise in the data follows Gaussian distribution, the least-squares function treats both the inliers and outliers in the data uniformly. This is undesirable because even one bad sample in the data can sway the optimisation to an unexpected convergence point. Therefore, outlier samples should be culled or down-weighted accordingly to maintain the optimisation and estimation process from getting influenced by them. Fortunately, in computer vision this has been long studied since the early 90s by Black \textit{et al.} \cite{Black:Anandan:1993} \cite{Black:Anandan:CVPR1991} and \cite{Black:etal:1998} who pioneered the use of robust cost functions, often termed M-estimators for estimating a statistically robust mean of the data. We adapted the standard $\mathcal{L}_2^{2}$ loss function with various popular M-estimators. The table below shows various M-estimators, $\rho$(x) and their corresponding derivatives, $\psi$(x).
\begin{center}
\begin{tabular}{| c | C{4cm} | C{3cm} |}
M-estimator       & $\rho$(x) & $\psi$(x) \\
\hline
& & \\
Huber \(\displaystyle \begin{cases}
  \text{if } |x| \leq \epsilon, \\
  \text{otherwise}.
\end{cases} \) & \(\displaystyle \begin{cases}
  \frac{x^2}{2}, \\
  \epsilon(|x| - \frac{\epsilon}{2})
\end{cases} \)  & \(\displaystyle \begin{cases}
  x, \\
  \epsilon\frac{x}{|x|}
\end{cases} \) \\ & & \\ 
Cauchy & $\frac{c^2}{2}\log(1+(\frac{x}{c})^2)$ & $\frac{x}{1+(\frac{x}{c})^2}$ \\& & \\
Geman-McClure & $\frac{x^{2}/2}{1+x^{2}}$ & $\frac{x}{(1+x^{2})^2}$ \\& & \\
Tukey\(\displaystyle \begin{cases} \text{if} |x| \leq c \\ \text{otherwise.} \end{cases} \)& \(\displaystyle \begin{cases}
  \frac{c^2}{6}(1-(1-(\frac{x}{c})^{2})^3)\\
  \frac{c^2}{6}
\end{cases} \) & \(\displaystyle \begin{cases}
  x(1-(\frac{x}{c})^2)^2, \\
  0
\end{cases} \)\\
\label{M estimators table}
\end{tabular}
\end{center}
The use of M-estimators has already started to trickle down in the deep learning community \textit{e.g.} Patraucean \textit{et al.} \cite{Patraucean:etal:ICLRW15} use a Huber loss function in the smoothness term to regularise the optic flow. We believe our library will also continue to encourage people to use different loss functions that are more pertinent to the tasks where Gaussian noise assumptions fall apart.  

\section{Application: Training on RGB-D Visual Odometry} 
We perform early experiments on visual odometry for both SO3 as well as SE3 motion that involves depth based warping. We believe this is the first attempt towards end-to-end system for Visual Odometry with deep learning. Since we are aligning images \textit{\`{a}} \textit{la} dense image registration methods, this allows us to do sanity checks on different layers \textit{e.g.} SE3 layer, 3D Grid Generator, and Projection layer all within the same network and optimisation scheme. Note that we could have also chosen to do minimisation on re-projection error of sparse keypoints as in classic Bundle Adjustment. However, this approach does not lend itself to generic iterative image alignment where each iteration provides a warped version of the reference image and can be fed back into the network for an end-to-end RNN based visual odometry system. Moreover, our approach is also naturally suited for unsupervised learning in the spirit of \cite{Patraucean:etal:ICLRW15} and \cite{Garg:etal:arXiv2016}.
\subsection{Network Architecture}
Our architecture is composed of a siamese network that takes in a pair of consecutive frames, $\mathcal{I}_{ref}$ and 
$\mathcal{I}_{live}$, captured at time instances $t$ and $t+1$ respectively, and returns a 6-DoF pose vector, $\delta_{pred}$ --- where the first three elements correspond to rotation and the last three to translation --- that transforms one image to the other. In case of pure rotation, the network predicts a 3$\times$1 vector. It is assumed that the scene is mostly static and rigid, and the motion perceived in the image is induced only via the camera movement. However, instead of na\"{i}vely comparing the predicted 6-DoF vector,  $\delta_{pred}$, with the corresponding ground truth vector, $\delta_{gt}$, we build upon the work of Patraucean \textit{et al.} \cite{Patraucean:etal:ICLRW15}, to warp the images directly using our customised \textit{3D Spatial Transformer} module, to compute the image alignment error as our cost function. This allows us to compare the transformations in the right space: na\"{i}ve comparison of 6-DoF vectors would have involved a tunable parameter beforehand to weigh the translation and rotation errors appropriately to define the cost function since they are two different entities. Searching for the right weighting can quickly become tedious and may not generalise well. Although \cite{Strobl:etal:IROS2006} are able to minimise a cost function by appropriately weighing the rotation and translation errors within optimal hand-eye coordination loop, this is not possible all the time. Discretising the poses as done in \cite{Agrawal:etal:ICCV2015} may hamper the accuracy of pose estimation. On the other hand, computing pixel error via warping, as often done in classic dense image alignment methods \cite{Lucas:Kanade:IJCAI1981},\cite{Drummond:Cipolla:CVPR1999}, allows to compare the transformations in the space of pixel intensities without having to tune any external parameters. Moreover, dense alignment methods have an added advantage of accurately recovering the transformations by minimising sum of squared differences of pixel values at corresponding locations in the two images \textit{i.e.}

\begin{eqnarray*}
\mathcal{C} &=& \frac{1}{2} \sum_{i=1}^{N}\bigg(\mathcal{I}_{ref}(\mathbf{x}) - \mathcal{I}_{live}(\pi(\mathsf{T}_{lr} \mathsf{\hat{p}}(\mathbf{x})))\bigg)^{2} \\
\end{eqnarray*}  

where $\mathbf{x}$ is a homogenised 2D pixel location in the reference image, $\mathsf{\hat{p}}(\mathbf{x})$ is the 4$\times$1 corresponding homogenised 3D point obtained by projecting the ray from that given pixel location $(x,y)$ into the 3D world via classic inverse camera projection and the depth, $\mathsf{d}(x,y)$, at that pixel location.

\begin{eqnarray}
\mathbf{x} = \left(
    \begin{array}{c}
      x \\
      y \\
      1
    \end{array}
  \right), 
&&
\mathsf{\hat{p}}(x,y) = \left(
    \begin{array}{c}
      \mathsf{K}^{-1}\mathbf{x} \cdot \mathsf{d}(x,y)\\
      1
    \end{array}
  \right) \\
\mathsf{K} = \begin{bmatrix} f_x & 0 & p_x \\ 0 & f_y & p_y \\ 0 & 0 & 1 \end{bmatrix} ,  
&&
\pi\left(
    \begin{array}{c}
      u \\
      v \\
      w
    \end{array}
  \right) 
 = 
\left(
    \begin{array}{c}
      f_x\frac{u}{w} + p_x \\
      \vspace{-2mm} \\
      f_y\frac{v}{w} + p_y
    \end{array}
  \right)   
\end{eqnarray}
$\mathsf{K}$ is the camera calibration matrix, $f_x$ and $f_y$ denote the focal lengths of the camera (in pixels) while $p_x$, $p_y$ are the coordinates of the camera center location. $\pi$ is the projection function that maps a 3D point to a 2D plane and $\mathsf{T}_{lr}$ (or $\mathsf{T}_{pred}$) is a 3$\times$4 matrix that transforms a 3D point in the reference frame to the live frame. In this work, we bridge the gap between learning and geometry based methods with our \textit{3D Spatial Transformer} module which explicitly defines these operations as layers that act as computational blocks with no learning parameters but allow backpropagation from the cost function to the input layers. 

\begin{figure}[h]
\centerline{
\includegraphics[width=\linewidth]{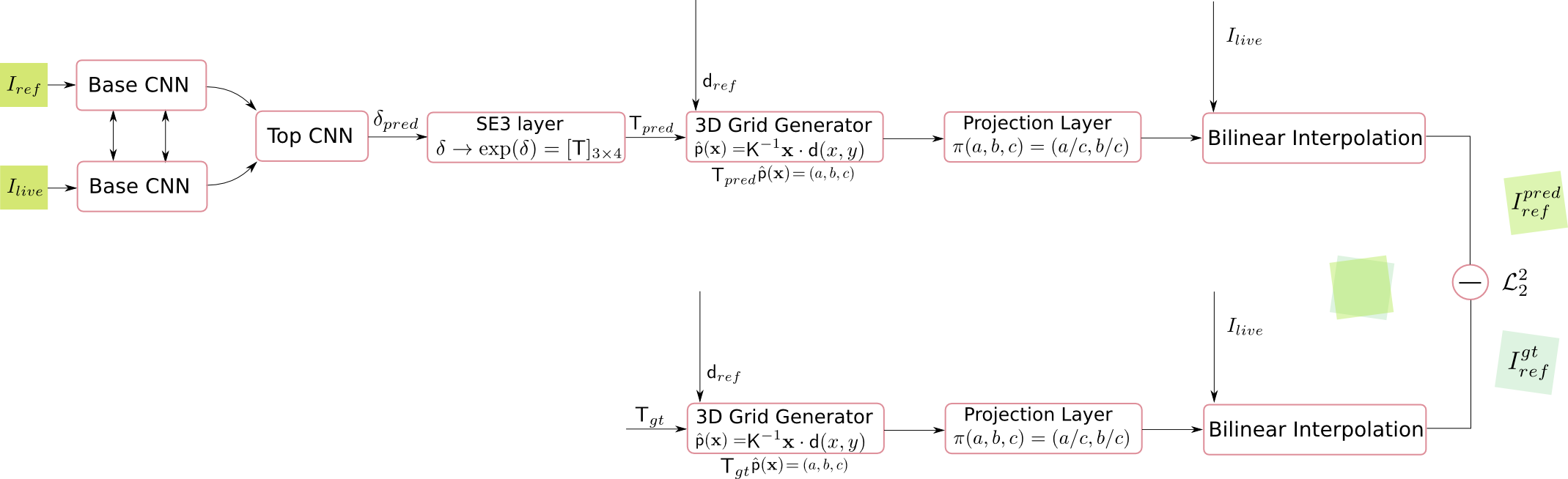}
}
\caption{\small
We train a siamese network to regress to the relative pose vector between the two consecutive frames, $I_{ref}$ and $I_{live}$. This pose vector is turned into a 3$\times$4 transformation matrix that transforms 3D points coming from the 3D grid generator and further projected into a 2D plane via projection layer which are used to generate a warped image. Additionally, the 3D grid generator needs an explicit depth-map as input to generate 3D points for any 3D warping.}  
\label{fig:STN SE3 arch}
\end{figure}

Figure \ref{fig:STN SE3 arch} shows an example of our customised STN for 3D transformation. The siamese network predicts a 6$\times$1 vector that is turned into a 3$\times$4 transformation matrix $\mathsf{T}_{pred}$ via SE3 layer. This matrix transforms the points generated by the 3D grid generator that additionally takes depth image as input and turns it into 3D points via inverse camera projection with $\mathsf{K}^{-1}$ as in Eq. 1. These transformed points are then projected back into the 2D image plane via the Projection layer (\textit{i.e.} the $\pi$ function) and further used to bilinearly interpolate the warped image as in the original STN \cite{Jaderberg:etal:NIPS2015}.

Our siamese network is inspired from the popular VGG-16 network \cite{Simonyan:Zisserman:ICLR2015} and uses 3$\times$3 convolutions in all but the last two layers where 2$\times$1 and 2$\times$2 convolutions are used to compensate for the 320$\times$240 resolution used as input as opposed to the 224$\times$224 used in original VGG-16. Figure \ref{fig:Siamese Network} shows our siamese network where two heads are fused early to ensure that the relevant spatial information is not lost by the depth of the network. We also avoid any pooling operations throughout the network, again to ensure that the spatial information is preserved. All convolutional layers, with the exception of the last three, are followed by a non-linearity. We found PReLUs \cite{He:etal:ICCV2015} to work better both in terms of convergence speed and accuracy than ReLUs for our network and therefore used them for all the experiments. We also experimented with recently introduced ELUs \cite{Clevert:etal:ICLR2016} but did not find any significant difference in the end to PReLUs. Weights of all convolution layers are initialised with MSRA initialisation proposed in \cite{He:etal:ICCV2015}. However, the last layer has the weights all initialised to zero. This is to ensure that the relative pose between the consecutive frames is initialised with Identity transformation, as commonly used in many dense image alignment methods.   

While one could use the pixel difference between the predicted live image, using the transformation returned by the siamese network, and the live image as the cost function, we chose instead to take the pixel difference between the predicted live image with the predicted transformation and the predicted live image with the ground truth transformation. This is because if there is significant motion between the input frames, warping may possibly lead to missing pixels in the predicted image which will get unnecessary penalised if compared against the live image directly since there is no explicit way to block out the corresponding pixels in the live image. However, if the predicted images from the predicted and ground truth transformations are compared, at optimal predicted transformations both should have the same missing pixels which would allow implicitly blocking out those pixels. Moreover, any external artefact in the images in the form of motion blur, intensity changes, or image noise would affect the registration since the cost function is a pixel-wise comparison. On the other hand, our way of comparing the pixels ensures that at convergence, the cost function is as close to zero as possible and is able to handle missing pixels appropriately. Ultimately, we only need a way to compare the predicted and ground truth transformations in the pixel space. We show early results of training on SO3 (pure rotation) and SE3 motion (involving rotation and translation).

\subsubsection{SO3 motion: pure rotation}
\begin{figure*}[hp]
\centerline{
\includegraphics[width=0.243\linewidth]{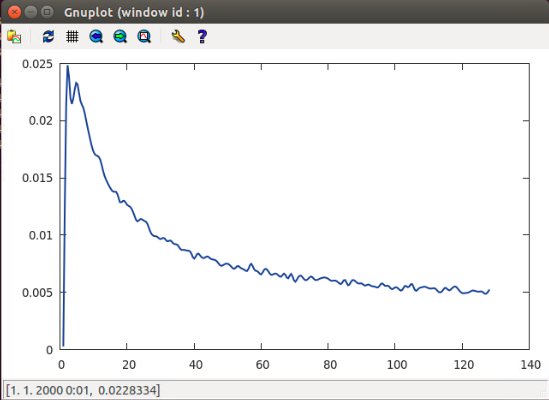}
\includegraphics[width=0.65\linewidth]{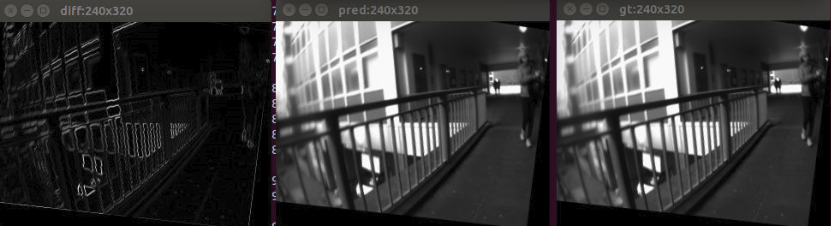}
}
\centerline{
\includegraphics[width=0.243\linewidth]{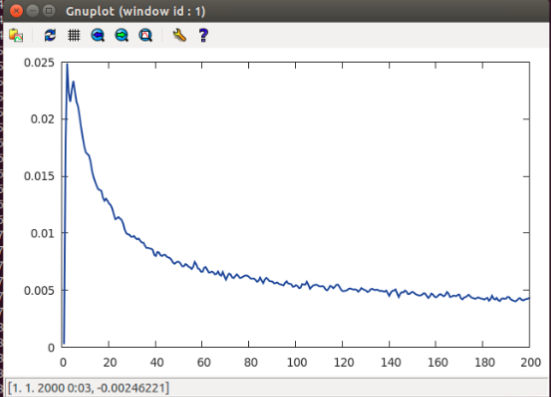}
\includegraphics[width=0.65\linewidth]{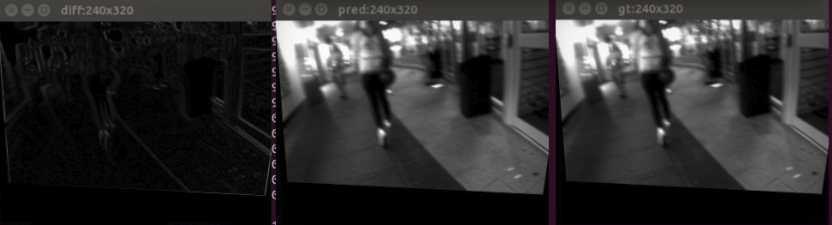}
}
\centerline{
\includegraphics[width=0.323\linewidth]{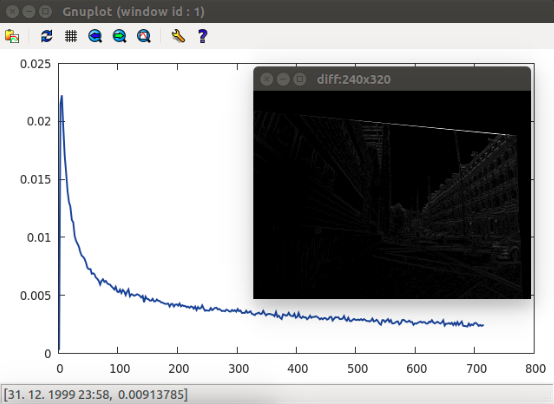}
\includegraphics[width=0.57\linewidth]{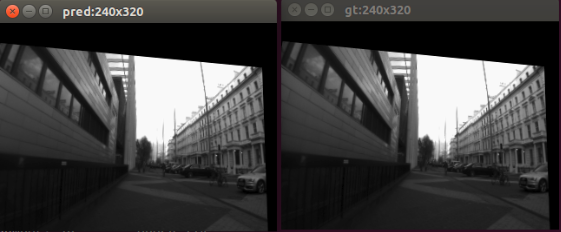}
}
\caption{\small
Training results on pure rotation motion. The graphs show how the training error decreases as number of epochs increase. This serves as a sanity check for our network that includes many new layers that we propose in this library. The improvement in the training is qualitatively evident from the difference images: early stages in the optimisation show high residual in the registration but as more epochs are thrown to the optimisation, the residual error gracefully decreases. } \vspace{1mm}
\label{fig:SO3 motion OKVIS}
\end{figure*}

To experiment with pure rotation motion, we gathered IMU readings of a camera undergoing rapid hand-held motion: we used \cite{Leutenegger:etal:IJRR2014} to capture an outdoor dataset but dropped the translation readings. This is only to ensure that the transformation in the images correspond to the real hand-held motion observed in real world. We use the rotation matrices to synthetically generate new images in the dataset and feed the corresponding pair through the network. We perform early experiments that serve as sanity checks for different layers working together in a network. Figure \ref{fig:SO3 motion OKVIS} shows how our system is able to register the images over a given training episode. The first row shows a high residual in the image registration but as the network improves with the training, the residual gradually starts to decrease: last row shows that the network is capable of registering images involving very large motion. Note that the prediction images at the start of training have no missing pixels (since the network is initialised with Identity transformation) but gradually start moving towards the ground truth image.

\subsubsection{SE3 motion: rotation and translation} 
\begin{figure*}[hp]
\centerline{
\includegraphics[width=0.125\linewidth]{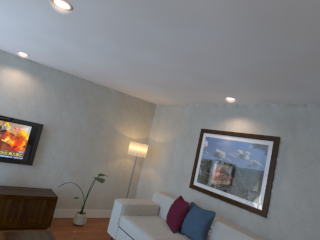}
\includegraphics[width=0.125\linewidth]{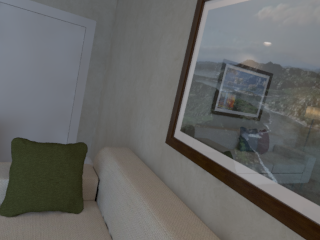}
\includegraphics[width=0.125\linewidth]{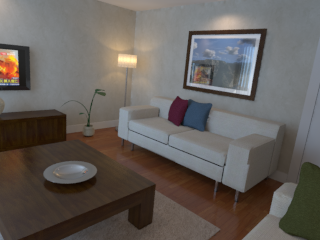}
\includegraphics[width=0.125\linewidth]{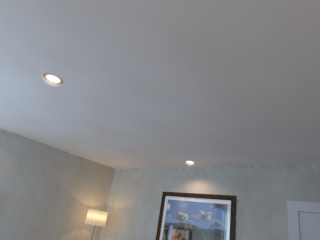}
\includegraphics[width=0.125\linewidth]{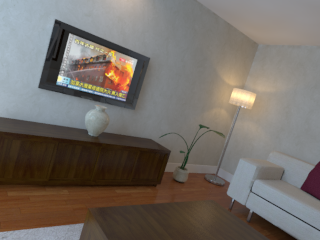}
\includegraphics[width=0.125\linewidth]{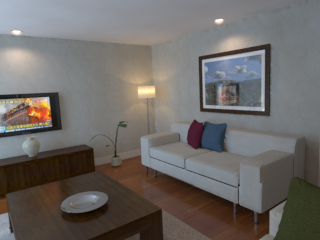}
\includegraphics[width=0.125\linewidth]{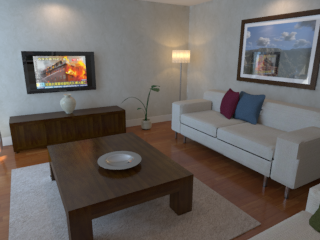}
}
\vspace{1mm}
\centerline{
\includegraphics[width=0.125\linewidth]{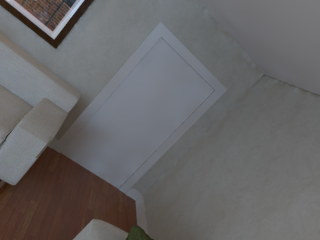}
\includegraphics[width=0.125\linewidth]{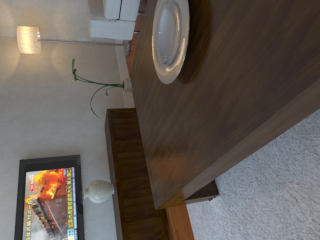}
\includegraphics[width=0.125\linewidth]{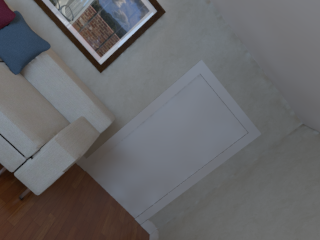}
\includegraphics[width=0.125\linewidth]{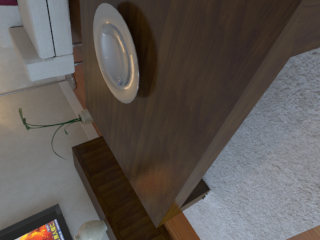}
\includegraphics[width=0.125\linewidth]{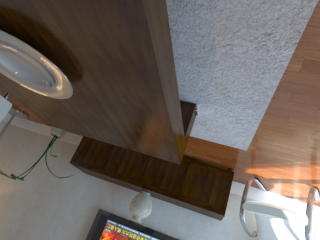}
\includegraphics[width=0.125\linewidth]{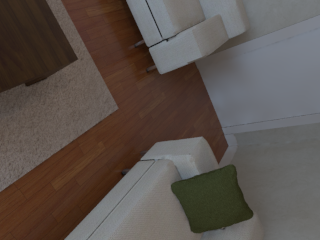}
\includegraphics[width=0.125\linewidth]{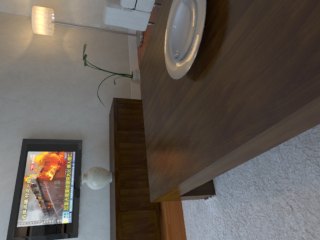}
}
\vspace{1mm}
\centerline{
\includegraphics[width=0.125\linewidth]{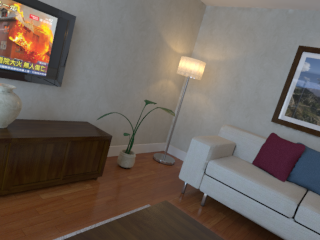}
\includegraphics[width=0.125\linewidth]{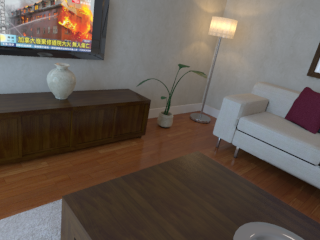}
\includegraphics[width=0.125\linewidth]{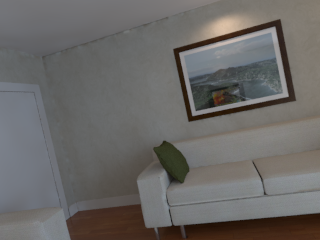}
\includegraphics[width=0.125\linewidth]{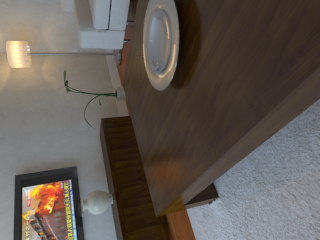}
\includegraphics[width=0.125\linewidth]{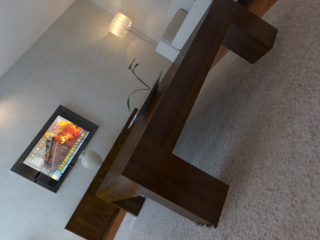}
\includegraphics[width=0.125\linewidth]{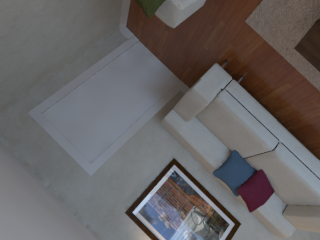}
\includegraphics[width=0.125\linewidth]{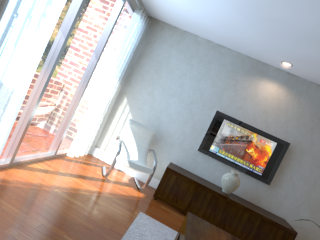}
}
\vspace{1mm}
\centerline{
\includegraphics[width=0.125\linewidth]{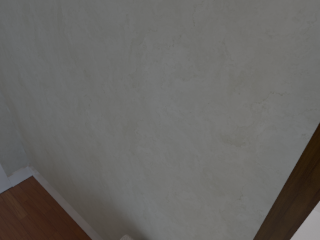}
\includegraphics[width=0.125\linewidth]{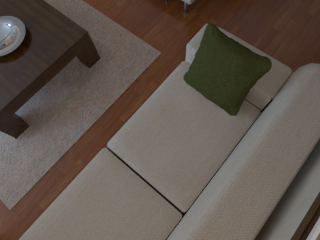}
\includegraphics[width=0.125\linewidth]{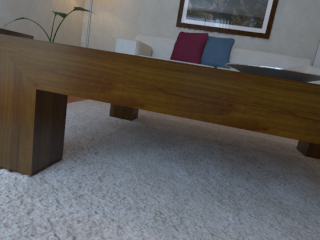}
\includegraphics[width=0.125\linewidth]{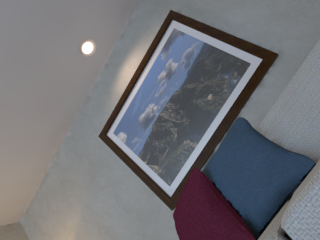}
\includegraphics[width=0.125\linewidth]{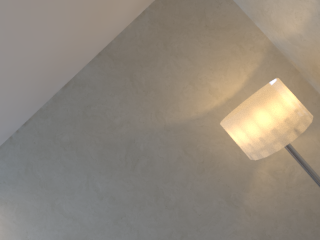}
\includegraphics[width=0.125\linewidth]{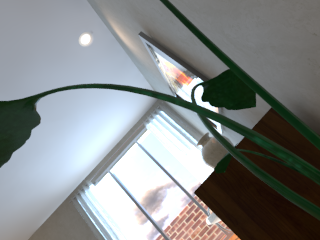}
\includegraphics[width=0.125\linewidth]{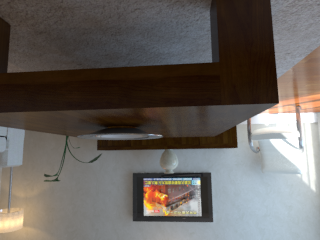}
}
\caption{\small
Sample frames from our new ICL-NUIM trajectory.} \vspace{1mm}
\label{fig:long traj ICL-NUIM}
\end{figure*}

SE3 motion needs depth to enable registration of two images involving both rotation and translation. This is possible with our SE3 layer that additionally takes in depth-map as input and produces the interpolation coordinates to be further used by the bilinear interpolation layer. We use ICL-NUIM \cite{Handa:etal:ICRA2014}
and generate a long trajectory of 9.5K frames and use this as our training set. Figure \ref{fig:long traj ICL-NUIM} shows samples of generated frames in this new trajectory. Since we need per-pixel depth for this experiment we opted for synthetic dataset only for convenience. In future, we would like to test our approach on real world data.

\begin{figure*}[htp]
\centerline{
\includegraphics[width=0.30\linewidth]{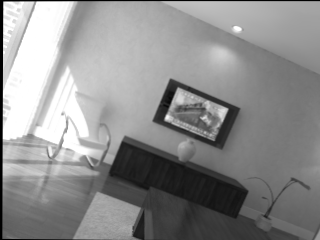}
\includegraphics[width=0.30\linewidth]{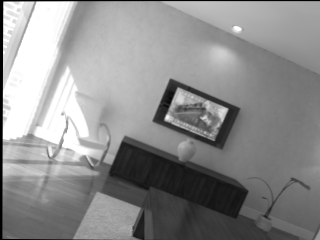}
\includegraphics[width=0.30\linewidth]{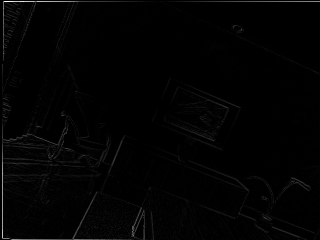}
}
\centerline{
\includegraphics[width=0.30\linewidth]{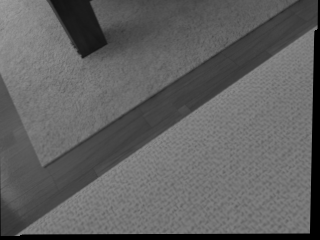}
\includegraphics[width=0.30\linewidth]{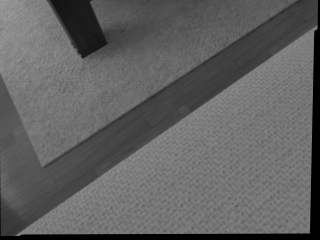}
\includegraphics[width=0.30\linewidth]{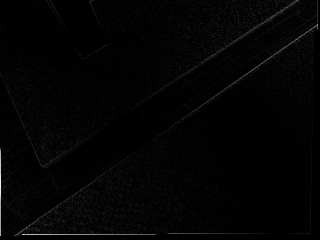}
}
\centerline{
\hfill
\makebox[0.50\linewidth][c] { \footnotesize{(a) Prediction }}
\hfill
\makebox[0.10\linewidth][c] { \footnotesize{(b) Ground Truth}}
\hfill
\makebox[0.50\linewidth][c] { \footnotesize{(c) Residual (difference)}}
}

\caption{\small Sample results on the new trajectory generated with ICL-NUIM dataset. The SE3 layer allows warping image with 3D motion and this is evident in the registration error in the residual image. Note that the relative motion between consecutive frames is generally slow in the whole trajectory. } \vspace{1mm}
\label{fig:SE3 motion ICL-NUIM}
\end{figure*}

Similar to the pure rotation (SO3) motion, we show early results on SE3 motion involving rotation and translation. Figure \ref{fig:SE3 motion ICL-NUIM} shows the network's ability to learn to align the predicted image with the ground truth image using depth that is given as an additional input to the 3D grid generator.

%RNN for Visual Odometry
%
%\begin{eqnarray*}
%h_{t+1} &=& f_{siamese}(\hat{\mathcal{I}}^{t+1}_{next}, {\mathcal{I}}_{next}) + h_{t} \\
%\hat{\mathcal{I}}^{t+1}_{next} &=& f_{warping}(\mathcal{I}_{cur}, h_{t+1})
%\end{eqnarray*}

%\cite{Srinivasan2001} \cite{Srinivasan1991}
%\section{Experiments}
%\begin{enumerate}
%\item Need to do experiments on TUM-RGBD dataset.
%\item Experiments on all 4 trajectories on ICL-NUIM.
%\end{enumerate}

\section{Future Work}
We have only shown training on visual odometry as sanity checks of our layers and their ability to blend in with the standard convolution neural network. In future, we would like to train both feed-forward as well as feedback connections based neural network on large training data. This data could either come from standard Structure from Motion
\cite{VisualSfM}, large scale synthetic datasets e.g. SceneNet \cite{Handa:etal:ARXIV2015} or large scale RGB or RGB-D videos for unsupervised learning. 

%\section{Things to add}
%\begin{enumerate}
%\item Dynamic Convolutions.
%\item FilterFlow.
%\item Unger Super-resolution.
%\item Adaptive Convolution Window, Pooling, per-pixel attention
%\item Talk about lens distortion models, 
%\end{enumerate}

\section{Conclusions}
We introduced a new library, \textbf{gvnn}, that allows implementation of various standard computer vision applications within a deep learning framework. In its current form, it allows end-to-end training for optic flow, disparity or depth estimation, visual odometry, small-scale bundle adjustment, super-resolution, place recognition with geometric invariance all with both supervised and unsupervised settings. In future, we plan to extend this library to include various different lens distortion models, camera projection models, IMU based transformation layers, sign distance functions, level-sets, and classic primal-dual methods \cite{Chambolle:2011} as RNN blocks to allow embedding higher order priors in the form of TGV \cite{Pock:etal:2011}. We hope that our library will encourage researchers to use and contribute towards making this a comprehensive and complete resource for geometric computer vision with deep learning in the same way the popular \textbf{rnn} package \cite{Leonard:etal:arXiv2015} has fostered research in recurrent neural networks in the community. Upon publication, we will release the full source code and sample application examples at \textbf{\url{https://github.com/ankurhanda/gvnn}}.

\section{Acknowledgements}
AH and AD would like to thank Dyson Technology Ltd. for kindly funding this research work.

%Talk about primal-dual, \textit{gvrnn}, talk about RNN based visual odometry, Bundle Adjustment, 

\bibliographystyle{splncs}
\bibliography{robotvision}

\end{document}